\documentclass[11pt]{article}
\usepackage{arabtex}
\usepackage{coling2018}
\usepackage{utf8}
\setcode{utf8}
\usepackage{url}
\usepackage{graphicx} 

\usepackage[stable]{footmisc}
\date{}

\begin{document}

\title{\textbf{Transliterating Kurdish texts in Latin into Persian-Arabic script}}

\author{
\begin{tabular}[t]{c}
Hossein Hassani\\
\textnormal{University of Kurdistan Hewl\^er}\\
\textnormal{Kurdistan Region - Iraq}\\
{\tt hosseinh@ukh.edu.krd}
\end{tabular}
}

\maketitle


\begin{abstract}
Kurdish is written in different scripts. The two most popular scripts are Latin and Persian-Arabic. However, not all Kurdish readers are familiar with both mentioned scripts that could be resolved by automatic transliterators. So far, the developed tools mostly transliterate Persian-Arabic scripts into Latin. We present a transliterator to transliterate Kurdish texts in Latin into Persian-Arabic script. We also discuss the issues that should be considered in the transliteration process. The tool is a part of Kurdish BLARK, and it is publicly available for non-commercial use\footnote{\url{https://kurdishblark.github.io/}}.
\end{abstract}

\section{Introduction}
\label{intro}

Kurdish is a multi-dialect that is written in different scripts \cite{hassani2016automatic}. The two most popular scripts are Latin and Persian-Arabic. However, not all Kurdish readers are familiar with both mentioned scripts that could be resolved by automatic transliterators. So far, the developed tools mostly transliterate Persian-Arabic scripts into Latin \cite{hassani2018blark,ahmadi2019rule}. We present a transliterator to transliterate Kurdish texts in Latin into Persian-Arabic script. Kurdish language processing requires endeavor by interested researchers and scholars to overcome the resource and tool scarcity to eliminate the obstacles in front of its tasks. The areas that need attention and the efforts required have been addressed in \cite{hassani2018blark}.

The rest of this paper is organized as follows. Section \ref{relatedwork} reviews the related work. Section \ref{translit} presents different parts of the dataset, such as the dictionary, phoneset, transcriptions, corpus, and language model. Finally, Section \ref{conclusion} concludes the paper and suggests some areas for future work.

\section{Related work}
\label{relatedwork}

Several scholars have addressed transliterators for Kurdish scripts \cite{esmaili2014towards,hassani2018blark,ahmadi2019rule}. Those studies mostly focused on transliteration from Persian-Arabic into Latin script. Particularly, \newcite{ahmadi2019rule} addressed several important issues in transliterating Kurdish texts in Persian-Arabic into Latin script and proposed and implemented appropriate resolutions for them.

Also, some online tools exist for transliteration of Kurdish scripts into each other, such as \url{https://www.lexilogos.com/keyboard/kurdish_conversion.htm} and \url{http://www.transliteration.kpr.eu/ku/en.html}. Those are mainly based on standard Latin scripts, and therefore, they miss some of the Persian-Arabic scripts such as {{\<ح>}, {\<ع>}, and {\<غ>}}. Other issues also exist in those transliterators, particularly when the text is in Kurmanji Kurdish. For example, both mentioned tools transliterate the Kurmuanji word "diînine" as ``{\RL{دیننه‌}}'', while the correct transliteration is ``{\RL{دئیننە}}''. 

Furthermore, to use the mentioned tools, the users must copy and paste their texts to the online tool that not only has limited space but also might not be desirable to the users from the copyright perspective. Our translator addresses those issues, and its script is publicly available under the GNU license. 

\section{The Transliterator}
\label{translit}
Our transliterator is a script in Python that could be used standalone. It receives an input file in the text that should be saved as UTF-8, and it provides the transliterated version of the text also in UTF-8. It resolves the mentioned issues that we addressed in Section \ref{relatedwork}. Particularly, it works for both Kurmanji and Sorani texts in Latin.

\begin{figure}[h!]
	\centering
	\includegraphics[width=1\linewidth]{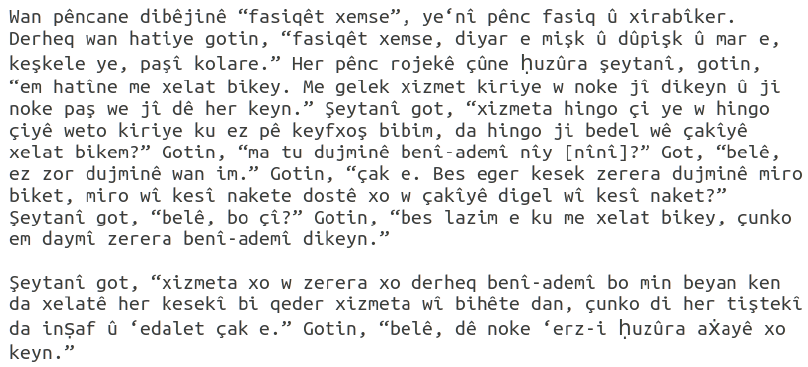}
	\caption{A Kurmanji text in Latin script.\protect\footnotemark}
	\label{fig:las}
\end{figure}

\begin{figure}[h!]
	\centering
	\includegraphics[width=1\linewidth]{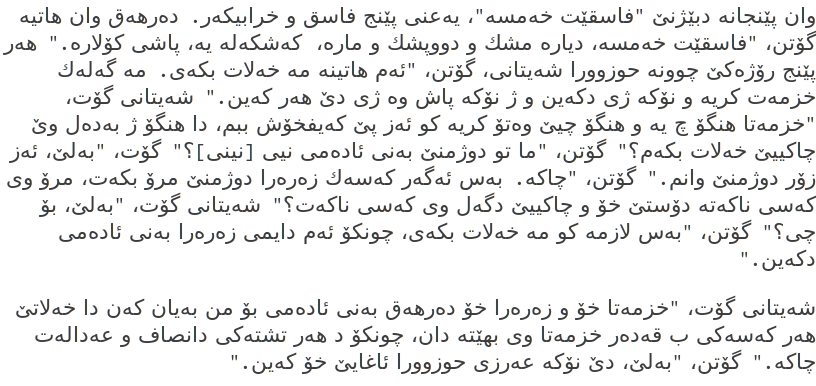}
	\caption{A Persian-Arabic version of the text in Figure \ref{fig:las} transliterated using our suggested script.}
	\label{fig:pas}
\end{figure}

Figure \ref{fig:las} shows a text in Latin script, and Figure \ref{fig:pas} illustrates its equivalent in Persian-Arabic script that our suggested script has transliterated.

\footnotetext{The text is from: {\"O}pengin, Ergin. 2021. Bazeber: Niv{\^i}sar{\^e}n Mela Se{'}{\^i}d \c{S}emd{\^i}nan{\^i} li ser çand {\^u} d{\^i}roka Kurdistana navend{\^i} [The texts of Mulla Said Shamdinani about the history and culture of Central Kurdistan]. Istanbul: Avesta.}

The transliterator is rule-based. We have recognized one-to-one one-to-many, two-to-one, two-to-many, and three-to-many letters transliterations. The \textit{many} equivalent is maximum three. The order that the transliterator applies those mapping is important. The mapping method and the mapping order were designed based on studying various texts and writing styles. For example, the case of  \textit{bizroke} that \cite{ahmadi2019rule} has addressed it transliteration from Persian-Arabic into Latin must also be considered in transliteration form Latin into Persian-Arabic. That is ``min'' must be written as ``{\RL{من}}''.  However, the users might still find issues when they transliterate using the tool. We would be grateful to report the possible issues to enhance the script.

The users of the transliterator may notice an issue with the full stop at the end of a paragraph when they use editors such as Microsoft Word or Libre Writer. When they open the text in those editors, the full stop at the end of the paragraphs might flip to the left. That case happens if the default direction for the input text in the editor is set to LTR. Using an appropriate command, according to the editor, the issue is resolved. For example, in Libre Writer, the user can select the entire text and then press {Ctrl+Righ Shift}. 

Also, it is usually necessary to change the font to the Unicode fonts, for example, UnikurdWeb, to view and print the output correctly.

\section{Conclusion}
\label{conclusion}
We presented a script that transliterates Kurdish texts in Latin script into Persian-Arabic texts. The script could be used standalone, and it is publicly available. The script resolves some issues that available online transliterators have not considered. As it is standalone, it could be used with requiring the Internet connection, and it does not have size limitations for its input document. In its current form, the transliterator provides proper output for Kurmanji and Sorani Kurdish.

In the future, we like to receive feedback~\footnote{Please kindly report any issues via Kurdish BLARK (\url{https://kurdishblark.github.io/})} from the transliterator concerning any possible issues to enhance the script. We also want to test it for Zazaki and Hawramit texts as they have special letters that are not used in Sorani and Kurmanji.

\section*{Acknowledgement} We are grateful to Dr. Ergin {\"O}pengin for his feedback and assistance in checking the accuracy of the transliterator.

\bibliographystyle{lrec}
\bibliography{l2pa}

\end{document}